\def\year{2022}\relax
\documentclass[letterpaper]{article} % DO NOT CHANGE THIS
\usepackage{aaai22}  % DO NOT CHANGE THIS
\usepackage{times}  % DO NOT CHANGE THIS
\usepackage{helvet}  % DO NOT CHANGE THIS
\usepackage{courier}  % DO NOT CHANGE THIS
\usepackage[hyphens]{url}  % DO NOT CHANGE THIS
\usepackage{graphicx} % DO NOT CHANGE THIS
\usepackage{float}
\usepackage{natbib}  % DO NOT CHANGE THIS AND DO NOT ADD ANY OPTIONS TO IT
\usepackage{caption} % DO NOT CHANGE THIS AND DO NOT ADD ANY OPTIONS TO IT
\DeclareCaptionStyle{ruled}{labelfont=normalfont,labelsep=colon,strut=off} % DO NOT CHANGE THIS
\usepackage[ruled,vlined,linesnumbered]{algorithm2e}
\usepackage{newfloat}
\usepackage{listings}
\floatstyle{ruled}
\newfloat{listing}{tb}{lst}{}
\floatname{listing}{Listing}
\usepackage[colorlinks=false]{hyperref}       % hyperlinks

\usepackage{amsmath}
\usepackage{bm}
\usepackage{multirow}
\usepackage{amssymb}
\usepackage{subfigure}
\usepackage{wrapfig}

\usepackage{graphicx}
\usepackage{bm}
\usepackage{multirow}
\usepackage{subfigure}
\usepackage{wrapfig}
\usepackage{url}
\usepackage{cleveref}
\usepackage{gensymb}
\usepackage{booktabs}

\title{Domain Adaptation by Maximizing Population Correlation with Neural Architecture Search}

\author{
    Zhixiong Yue, Pengxin Guo, and Yu Zhang    
}

\affiliations{
Department of Computer Science and Engineering, Southern University of Science and Technology\\
11860002@mail.sustech.edu.cn,12032913@mail.sustech.edu.cn,yu.zhang.ust@gmail.com
}

\begin{document}

\maketitle

\begin{abstract}

In Domain Adaptation (DA), where the feature distributions of the source and target domains are different, various distance-based methods have been proposed to minimize the discrepancy between the source and target domains to handle the domain shift. In this paper, we propose a new similarity function, which is called Population Correlation (PC), to measure the domain discrepancy for DA. Base on the PC function, we propose a new method called Domain Adaptation by Maximizing Population Correlation (DAMPC) to learn a domain-invariant feature representation for DA. Moreover, most existing DA methods use hand-crafted bottleneck networks, which may limit the capacity and flexibility of the corresponding model. Therefore, we further propose a method called DAMPC with Neural Architecture Search (DAMPC-NAS) to search the optimal network architecture for DAMPC. Experiments on several benchmark datasets, including Office-31, Office-Home, and VisDA-2017, show that the proposed DAMPC-NAS method achieves better results than state-of-the-art DA methods.

% These methods rely on the domain-specific latent feature representations extracted from the bottleneck layer after the pre-trained backbone. However, the neural architecture of bottleneck layer is yet hand-crafted and can be improved by a better architecture design. To tackle these challenges, we propose a novel distance based method for UDA called Domain Adaptation by Maximizing Population Correlation (DAMPC). We further design a framework DAMPC-NAS that employ DAMPC with Neural Architecture Search (NAS) method to find optimal architecture for feature fusion between source and target domain. Experiments on Office-31, Office-Home, and VisDA-2017 show that DAMPC-NAS  can achieve better results than other distance based methods in UDA.

\end{abstract}

\section{Introduction}
With access to large-scale labeled data, deep neural networks have achieved state-of-the-art performance among a variety of machine learning problems and applications \cite{krizhevsky2012imagenet, oquab2014learning, donahue2014decaf, yosinski2014transferable, ren2015faster, he2016deep, he2017mask}. However, with intolerably time-consuming and labor-expensive costs, it is hard for a target domain of interest to collect enough labeled data for model training. One solution is to transfer a deep neural network trained on a data-sufficient source domain to the target domain where only unlabeled data is available. However, this learning paradigm suffers from the shift in data distributions across different domains, which brings a major obstacle in adapting predictive models for the target task.

Domain Adaptation (DA) \cite{pan2009survey, yang2020transfer}
aims to learn a high-performance learner on a target domain via utilizing the knowledge transferred from a source domain, which has a different but related data distribution to the target domain. A number of DA methods aim to bridge the gap between source and target domains so that the classifier learned in the source domain can be applied to the target domain. To achieve this goal, recent DA works can be grouped into two main categories: \textit{distance-based} methods \cite{ben2007analysis, ben2010theory, zhuang2015supervised, tzeng2014deep, long2015learning, courty2016optimal, sun2016return, sun2016deep, zellinger2017central, chen2019joint} and  \textit{adversarial} DA methods \cite{ganin2016domain, long2017conditional, pei2018multi, tzeng2017adversarial, saito2018maximum}. Both categories aim to learn the domain-invariant feature representations. In this paper, we mainly focus on distance-based DA methods.

For distance functions adopted by DA, the first attempt is the \textit{Proxy} $\mathcal{A}$-\textit{distance}
%, given by $\hat{d}_{\mathcal{A}}=2(1-2 \epsilon)$, where $\epsilon$ is the generalization error on the problem of discriminating between source and target samples
\cite{ben2010theory}, which aims to minimize the generalization error by discriminating between source and target samples. \textit{Maximum Mean Discrepancy} (MMD) \cite{gretton2006kernel} is a popular distance measures between two domains and it has been used in Deep Domain Confusion (DDC) \cite{tzeng2014deep} and Deep Adaptation Network (DAN) \cite{long2015learning}.
% For other distance based methods, Transfer Learning with Deep Autoconders (TLDA) \cite{zhuang2015supervised} utilizes the \textit{Kullback-Leibler divergence}. CORrelation ALignment (CORAL) \cite{sun2016return, sun2016deep} adopts the \textit{second-order statistics}, and Central Moment Discrepancy (CMD) \cite{zellinger2017central} proposes a new distance measurement metric.
Although numerous distance-based DA methods have been proposed, learning the domain-invariant feature representation is still challenging since distances in a high-dimensional space may be difficult to truly reflect the domain discrepancy. Moreover, all of these methods are developed by using hand-crafted network architectures. Since the difficulty levels of different DA tasks are not the same, accomplishing complex tasks may require a more sophisticated network architecture than easy tasks, hence, using the same hand-crafted network architecture may limit the capacity and versatility of DA methods.

To alleviate these limitations, in this paper, we propose a new similarity function, which is called Population Correlation (PC), to measure the similarity between the source and target domains. Based on the PC function, we propose a novel domain adaptation method called Domain Adaptation by Maximizing Population Correlation (DAMPC). DAMPC aims to maximize the PC between the source and target domains so that a learning model can learn a domain-invariant feature representation. Specifically, With the PC defined as the maximum of pairwise correlations between source and target samples, the proposed DAMPC method maximize it to force the two domains to have similar distributions as well as minimizing the classification loss on the labeled source samples. Built on the DAMPC method, we design a reinforcement-based Neural Architecture Search (NAS) method called DAMPC-NAS to search an optimal network architecture for DAMPC. In this way, DAMPC-NAS can learn suitable network architectures for different DA tasks.
%$\mathrm{corr}\left(F(\mathbf{x}^s),F(\mathbf{x}^t)\right)$ where $F(\cdot)$ denotes the feature extraction network and $\mathrm{corr}(\mathbf{x}_1,\mathbf{x}_2)=\frac{\mathbf{x}_1^T\mathbf{x}_2}{\|\mathbf{x}_1\|_2\|\mathbf{x}_2\|_2}$ denotes the correlation between two vectors.
%Our proposed DAMPC approach is similar to DDC, DAN, and D-CORAL in the sense that a new loss term (i.e., PC) is added to decrease the domain shift between two domains.
To the best of our knowledge, the proposed DAMPC-NAS method is the first NAS framework designed for similarity-based DA methods. DAMPC-NAS is also one of few works that integrate NAS methods into deep DA methods. Our contributions are summarized as follows.
\begin{itemize}
    \item We propose a new similarity measure, i.e., PC, to measure the domain similarity. Based on the PC, we propose a DAMPC method for DA.
    \item We design the DAMPC-NAS framework to search optimal network architectures for the proposed DAMPC method.
    % To the best of our knowledge, we are the first to use NAS to search optimal network architecture for different distance based methods.
    \item Experimental results on three benchmark datasets demonstrate the effectiveness of the proposed methods.
\end{itemize}

\section{Related Work}

\paragraph{Neural Architecture Search}
NAS aims to design the architecture of a neural network in an automated way. Comparing with manually designed architectures of neural networks, NAS has demonstrated the capability to find architectures with state-of-the-art performance in various tasks \cite{pham2018efficient, lsy19, ghiasi2019fpn}. For example, the NAS-FPN method \cite{ghiasi2019fpn} leverages NAS to learn an effective architecture of the feature pyramid network for object detection.

Although NAS can achieve satisfactory performance, the high computational cost of the searching procedure makes NAS less attractive. To accelerate the search procedure, one-shot NAS leverages a supergraph, which contains all the candidate architectures in the search space. In the supergraph, weights of operations on edges are shared across different candidate architectures. ENAS \cite{pham2018efficient} employs a reinforcement-based method to train a controller that samples architectures from a supergraph with a weight sharing mechanism. DARTS \cite{lsy19} search architectures with a differentiable objective function based on a supergraph that uses the softmax function to contain all candidate operations on each edge. The final architecture is determined based on the weights corresponding to the candidate operations on each edge.

\paragraph{Domain Adaptation} DA aims to transfer the knowledge learned from a source domain with labeled data to a target domain without labeled data, where there is a domain shift between domains. As discussed in the introduction, recent works in DA can be mainly grouped into two categories: distance-based methods and  adversarial DA methods. In this paper, we mainly focus on \textit{distance-based} methods, which minimize the discrepancy between the source and target domains via some measures, including the MMD used in DDC \cite{tzeng2014deep}, DAN \cite{long2015learning}, Weighted Domain Adaptation Network (WDAN) \cite{yan2017mind}, Joint Adaptation Networks (JAN) \cite{long2017deep}, and Deep Subdomain Adaptation Network (DSAN) \cite{zhu2020deep}, the \textit{Kullback-Leibler divergence} adopted in Transfer Learning with Deep Autoconders (TLDA) \cite{zhuang2015supervised}, the \textit{second-order statistics} utilized in CORrelation ALignment (CORAL) \cite{sun2016return, sun2016deep}, and the Central Moment Discrepancy (CMD) \cite{zellinger2017central}.

\begin{figure*}[htbp]
\centering
\includegraphics[width=0.95\linewidth]{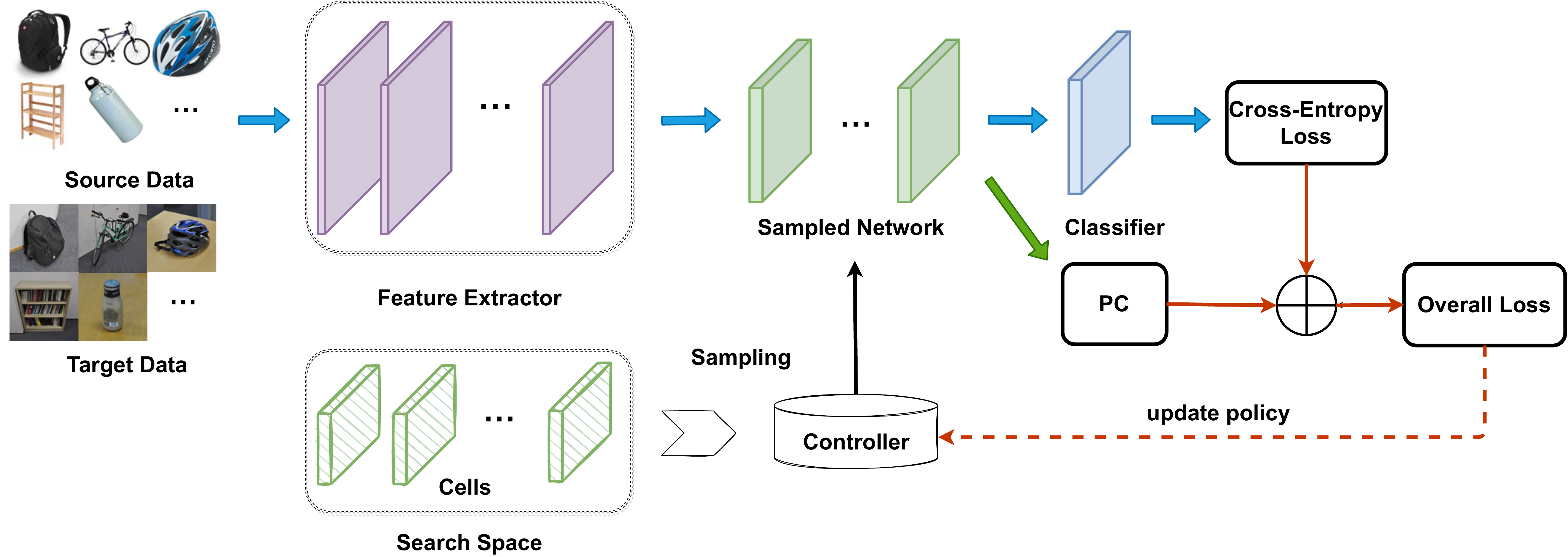}
\caption{Overview of the DAMPC-NAS framework. Source and target data first go through the feature extractor to extract hidden features. The controller samples cell choices for each cell and connections between the cells from search space to generate the architecture of the sampled network. Source and target data with the extracted feature representation then go through the sampled network. Finally, the cross-entropy loss is minimized and the PC is maximized. The controller's policy is updated by the reward of the negative overall loss.}
\label{fig:model}
\end{figure*}

\paragraph{Neural Architecture Search for Domain Adaptation} There are few works on NAS for DA. To improve the generalization ability of neural networks for DA, \citet{li2020adapting} analyze the generalization bound of neural architectures and propose the AdaptNAS method to adapt neural architectures between domains.  \citet{li2020network} propose a DARTS-like method for DA, which combines DARTS and DA into one framework. \citet{robbiano2021adversarial} aim to learn a auxiliary branch network from data for an adversarial DA method. In this paper, different from those works, we aim to leverage NAS to search optimal neural architectures for the proposed DAMPC method.

\section{Methodology}

In this section, we introduce the proposed PC similarity and the DAMPC method as well as the DAMPC-NAS method.

%In the era of deep neural networks, the main adaption idea is to learn domain-invariant features by minimizing difference between source and target feature distributions in an end-to-end way.
%To achieve this goal, we propose a new method DAMPC that maximizes the PC between the source and target feature representations. Furthermore, we design the DAMPC-NAS framework that leverages Neural Architecture Search method to search optimal network architecture for DAMPC.

%Figure \ref{fig:model} shows a sample our proposed DAMPC-NAS architecture.

\subsection{Population Correlation}

We first present the definition of PC. Here we study DA under the unsupervised setting. That is, the target domain has unlabeled data only. In DA, the source domain $\mathcal{D}_{s}=\left\{\left(\mathbf{x}_{i}^{s}, \mathbf{y}_{i}^{s}\right)\right\}_{i=1}^{n_{s}}$ has $n_s$ labeled samples and the target domain $\mathcal{D}_{t}=\left\{\mathbf{x}_{j}^{t}\right\}_{j=1}^{n_{t}}$ has $n_t$ unlabeled samples. To adapt the classifier trained on the source domain to the target domain, one solution is to minimize the domain discrepancy or equivalently maximize the domain similarity. To achieve this, we propose the PC to measure the similarity between the source and target domains. Specifically, suppose $F(\cdot)$ is the feature extraction network. Then the PC between the source and target domains can be computed based on each pair of source and target samples as
\begin{equation} \label{eq:pc_loss}
\begin{aligned}
\mathrm{PC}(\mathcal{D}^s,\mathcal{D}^t)=&\frac{1}{n_s}\sum_{i=1}^{n_s}\max_{j\in[n_t]}\mathrm{corr}\left(F(\mathbf{x}^s_i),F(\mathbf{x}^t_j)\right)\\
&+\frac{1}{n_t}\sum_{j=1}^{n_t}\max_{i\in[n_s]}\mathrm{corr}\left(F(\mathbf{x}^s_i),F(\mathbf{x}^t_j)\right),
\end{aligned}
\end{equation}
where $\|\cdot\|_2$ denotes the $\ell_2$ norm of a vector, $\mathrm{corr}(\mathbf{x}_1,\mathbf{x}_2)=\frac{\mathbf{x}_1^T\mathbf{x}_2}{\|\mathbf{x}_1\|_2\|\mathbf{x}_2\|_2}$ denotes the correlation between two vectors, and $[n]$ denotes a set of integers $\{1,\ldots,n\}$ for an integer $n$. Here we use the cosine similarity to calculate the correlation between two vectors, thus the larger the PC value is, the more similar the two domains are.

% Note that the value range of the PC is $-2$ to $2$. When the PC equals the minimum value $-2$, the target domain and the source domain are not similar at all. When the PC equals maximize value $2$, the target and source domain are highly correlated.

\subsection{DAMPC}

Built on the PC introduced in the previous section, in this section, we present the proposed DAMPC method which aims to learn a domain-invariant feature representation. For DA tasks, the hidden feature representations learned by the feature extraction network should be not only discriminative to train a strong classifier but also domain-invariant to both the source and target domains. Only maximizing the PC can help learn a domain-invariant feature representation and % learning degenerated feature representations, in which the learned feature representations are domain-invariant but not discriminative. Conversely,
only minimizing the classification loss is to learn a discriminative feature representation. Therefore, we combine the classification loss and the PC to obtain the final objective function, which is formulated as
\begin{equation} \label{eq:obj}
\mathcal{L}_{\text{DAMPC}} = \frac{1}{n_s}\sum_{i=1}^{n_s}l(C(F(\mathbf{x}^s_i)),y^s_i)-\lambda\mathrm{PC}(\mathcal{D}^s,\mathcal{D}^t),
\end{equation}
where $\lambda$ is a trade-off parameter, $C(\cdot)$ denotes the classification layer, and $l(\cdot,\cdot)$ denotes the classification loss such as the cross-entropy loss.

By minimizing Eq. (\ref{eq:obj}), the final learned feature representations are not only discriminative for classification but also domain-invariant for the adaptation.

\subsection{DAMPC-NAS}

In this section, we introduce the proposed DAMPC-NAS framework that finds an optimal architecture for the DAMPC method introduced in the previous section. An overview of the DAMPC-NAS framework is shown in Figure \ref{fig:model}.

\subsubsection{Cell-based Search Space}

\begin{figure}[htbp!]
\centering
\includegraphics[width=0.95\linewidth]{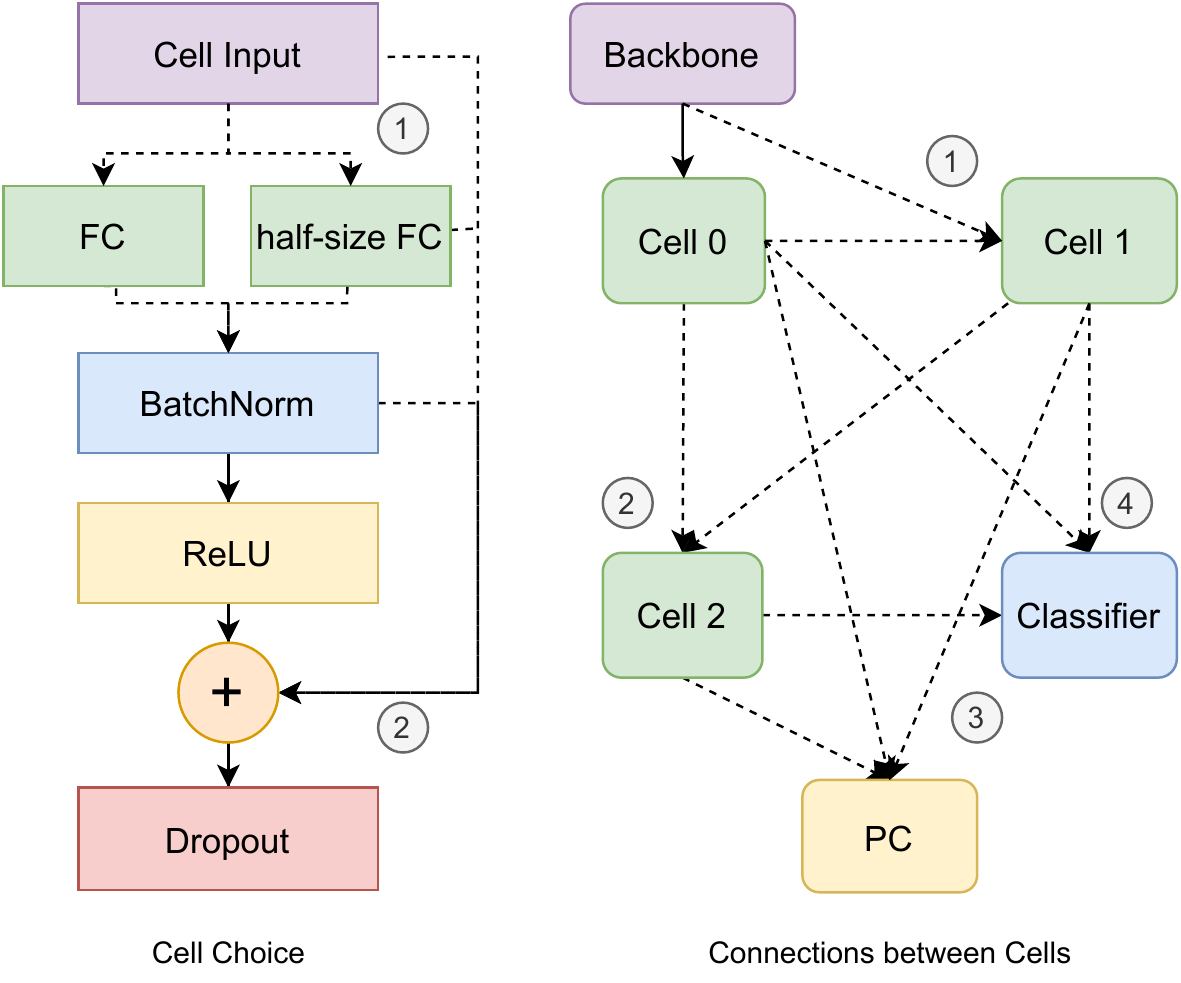}
\caption{The search space of the DAMPC-NAS method. Dashed lines represent possible search choices and numbered grey circles indicate the order of choices generated from the controller.}
\label{fig:search_space}
\end{figure}

We design the search space on the top of the Resnet-50 backbone, whose architecture is kept fixed, and hence we only search the architecture after the backbone. The search space of the DAMPC-NAS method consists of two parts: within cells and between cells. We design the cell as the composition of the fully connected layer, batch-norm layer, and dropout layer as well as the associated activation functions. Within the cell, we search for the size of the fully connected layer and the location of the skip connection. Specifically, the search choice of the fully connected layer in a cell can be `the same as input size' or `the half of input size'. The starting location of the skip connection can be chosen from the cell input, the fully connected layer, and the batch-norm layer. Between the cells, we search for input and output connections of the $N$ cells. For example, if there are three cells in the search space, i.e., $N=3$, the input of ``Cell 1'' can be chosen from the outputs of ``Backbone'' and ``Cell 0'', and the input of ``Cell 2'' can be chosen from the outputs of ``Cell 0'' and ``Cell 1'', hence the input of a cell can be chosen from the outputs of the previous two cells. The calculation of PC can be choose from one of outputs of all cells. Moreover, One of the outputs from the $N$ cells, i.e., ``Cell 0'', ``Cell 1'' and ``Cell 2'', can connect to the classifier trained on source domain data. Hence, the total search space has $(2\times3)^N2^{N-1}N^2$ configurations.  An illustration of the search space in the DAMPC-NAS method is shown in Figure \ref{fig:search_space}. In experiments, for efficiency, we use the search space with $N=3$ cells for all experiments.

\subsubsection{Searching optimal architecture}

The searching algorithm for the DAMPC-NAS method is described in Algorithm \ref{alg:nas}. DAMPC-NAS is a reinforcement-based NAS framework which leverages a controller network to sample architectures from the search space. The controller network is a LSTM that samples search choice via a softmax classifier. We denote by $\theta$ the learnable parameters of the controller. The policy of the controller is denoted by $\pi(m;\theta)$.

In each epoch, the training procedure of DAMPC-NAS consists of two phases. In the first phase, we fix parameters of the controller $\theta$ and train the shared weights $\omega$ in the search space $\mathcal{A}_{space}$. Specifically, the controller samples an architecture $\mathcal{A}_{m}$ from the search space $\mathcal{A}_{space}$ with policy $\pi(m;\theta)$. For each mini-batch from $\mathcal{D}_{s}$ and $\mathcal{D}_{t}$, $\mathcal{L}_{\text{DAMPC}}$ is computed according to Eq. (\ref{eq:obj}) and the shared weights $\omega_{m}$ of the sampled architecture are updated by minimizing $\mathcal{L}_\text{DAMPC}$. In the second phase, we fix all the shared weights $\omega$ in the search space $\mathcal{A}_{space}$ and update the parameter $\theta$ of the controller. Specifically, after one epoch of training, $-\mathcal{L}_\text{DAMPC}$ is used as the reward to update the policy $\pi(m;\theta)$ in the controller. The gradient is computed via the REINFORCE algorithm \cite{williams1992simple} with a moving average baseline.

\begin{algorithm}
\SetAlgoLined
\SetKwInOut{Input}{Input}
\SetKwInOut{Output}{Output}
\SetKwInOut{Return}{Return}
\SetKwComment{Comment}{// }{}
\SetCommentSty{normalsize}
\Input{source data $\mathcal{D}_{s}$, target data $\mathcal{D}_{t}$, the number of training epochs $n_{epochs}$}
\Output{The searched architecture with learned weights}
 initialize controller\;
 \For{$i\gets0$ \KwTo $n_{epochs}$}{
 sample $\mathcal{A}_{m}$ from $\mathcal{A}_{space}$ with policy $\pi(m;\theta)$\;
 \Comment{fix controller policy $\pi(m;\theta)$ and train $\omega$ in $\mathcal{A}_{space}$}
  \For{\text{mini-batch} in $\mathcal{D}_{s}$ and $\mathcal{D}_{t}$}{
    compute $\mathcal{L}_\text{DAMPC}$ in Eq. (\ref{eq:obj}) with $\mathcal{A}_{m}$\;
    % generate pseudo-label from target prediction\;
    update $\omega_m$ in $\mathcal{A}_{space}$\ with $\mathcal{L}_\text{DAMPC}$\;
    }
 \Comment{fix $\omega$ in $\mathcal{A}_{space}$ and update $\theta$ in policy $\pi(m;\theta)$}
    calculate reward of $\mathcal{A}_{m}$ as $R_m=-\mathcal{L}_{\text{DAMPC}}$\;
    update $\theta$ in $\pi(m;\theta)$ with reward $R_m$\;
 }
 \Return{$\mathcal{A}_{m}$ with trained weights $\omega_m$}
 \caption{Overview of DAMPC-NAS}
 \label{alg:nas}
\end{algorithm}

In summary, the DAMPC-NAS method is a one-shot style NAS method. That is, the DAMPC-NAS method trains a supernet that contains all shared parameters in the search space during the searching process. The DAMPC-NAS method samples a child network in each epoch to calculate the loss function defined in Eq. (\ref{eq:obj}) and updates its shared parameters in the search space. Parameters in the controller are updated by the reward, which is the negative loss of the sampled child network. After searching, all weights of the final architecture are retained for testing. Different from two-stage one-shot NAS methods, there is no need for the DAMPC-NAS method to retrain the final architecture from scratch for testing since DAMPC-NAS can directly optimize the objective in Eq. (\ref{eq:obj}), which is just the negative reward for the controller, in an end-to-end manner. In this way, the architecture is optimized alongside child networks' parameters. Therefore, the final architecture derived from the DAMPC-NAS method can be deployed directly without parameter retraining, which improves the efficiency.

\section{Experiments}

In this section, we empirically evaluate the proposed method.

\begin{table*}[htbp]
\centering
\caption{Accuracy (\%) on the Office-31 dataset with ResNet-50 as the backbone.}
% \resizebox{\linewidth}{!}{
\begin{tabular}{llccccccc}
\toprule
 Type & Method & A$\rightarrow$D & A$\rightarrow$W & D$\rightarrow$A & D$\rightarrow$W & W$\rightarrow$A & W$\rightarrow$D & Avg \\
 \midrule
  & ResNet-50 \cite{he2016deep} & 68.9 & 68.4 & 62.5 & 96.7 & 60.7 & 99.3 & 76.1 \\
 \cmidrule(){1-9}
 \multirow{6}*{Dist Based} & JDA \cite{long2013transfer} & 80.7 & 73.6 & 64.7 & 96.5 & 63.1 & 98.6 & 79.5  \\
 & DDC \cite{tzeng2014deep} & 76.5 & 75.6 & 62.2 & 96.0 & 61.5 & 98.2 & 78.3 \\
 & DAN \cite{long2015learning} & 78.6 & 80.5 & 63.6 & 97.1 & 62.8 & 99.6 & 80.4 \\
 & D-CORAL \cite{sun2016deep} & 81.5 & 77.0 & 65.9 & 97.1 & 64.3 & 99.6 & 80.9 \\
 & JAN \cite{long2017deep} & 84.7 & 85.4 & 68.6 & 97.4 & 70.0 & 99.8 & 84.3  \\
 & MDDA \cite{wang2020transfer} & 86.3 & 86.0 & \textbf{72.1} & 97.1 & \textbf{73.2} & 99.2 & 85.7 \\
 \cmidrule(){1-9}
 \multirow{4}*{Adv Based} & DANN \cite{ganin2015unsupervised} & 79.7 & 82.0 & 68.2 & 96.9 & 67.4 & 99.1 & 82.2 \\
 & ADDA \cite{tzeng2017adversarial} & 77.8 & 86.2 & 69.5 & 96.2 & 68.9 & 98.4 & 82.9 \\
 & CAN \cite{zhang2018collaborative} & 85.5 & 81.5 & 65.9 & 98.2 & 63.4 & 99.7 & 82.4 \\
 & DDAN \cite{wang2020transfer} & 84.9 & 88.8 & 65.3 & 96.7 & 65.0 & \textbf{100.0} & 83.5 \\
 \midrule
%  & DAMPC (Ours) & \textbf{88.35} & \textbf{91.32}  & 70.36 & \textbf{98.49} & 69.05 & \textbf{100.0} & \textbf{86.26} \\
 & DAMPC-NAS (Ours) & \textbf{89.16} & \textbf{93.08}  & 70.36 & \textbf{98.74} & 69.05 & \textbf{100.0} & \textbf{86.69} \\
 \bottomrule
 \end{tabular}
%  }
 \label{tab:office31}
\end{table*}

\begin{table*}[ht]
%\vskip -0.1in
\centering
%\vspace{0.2cm}
\caption{Accuracy (\%) on the Office-Home dataset with ResNet-50 as the backbone.}
\resizebox{\linewidth}{!}{
\begin{tabular}{llccccccccccccc}
\toprule
 Type & Method & Ar$\rightarrow$Cl & Ar$\rightarrow$Pr & Ar$\rightarrow$Rw & Cl$\rightarrow$Ar & Cl$\rightarrow$Pr & Cl$\rightarrow$Rw & Pr$\rightarrow$Ar & Pr$\rightarrow$Cl & Pr$\rightarrow$Rw & Rw$\rightarrow$Ar & Rw$\rightarrow$Cl & Rw$\rightarrow$Pr & Avg  \\
 \midrule
&  ResNet-50 \cite{he2016deep} & 34.9 & 50.0 & 58.0 & 37.4 & 41.9 & 46.2 & 38.5 & 31.2 & 60.4 & 53.9 & 41.2 & 59.9 & 46.1 \\
\cmidrule(){1-15}
 \multirow{4}*{Dist Based} & JDA  \cite{long2013transfer} & 38.9 & 54.8 & 58.2 & 36.2 & 53.1 & 50.2 & 42.1 & 38.2 & 63.1 & 50.2 & 44.0 & 68.2 & 49.8 \\
%  & DDC \cite{tzeng2014deep} & \\
 & DAN \cite{long2015learning} & 43.6 & 57.0 & 67.9 & 45.8 & 56.5 & 60.4 & 44.0 & 43.6 & 67.7 & 63.1 & 51.5 & 74.3 & 56.3 \\
 & D-CORAL \cite{sun2016deep} & 42.2 & 59.1 & 64.9 & 46.4 & 56.3 & 58.3 &  45.4 & 41.2 & 68.5 & 60.1 & 48.2 & 73.1 & 55.3 \\
 & JAN \cite{long2017deep} & 45.9 & 61.2 & 68.9 & 50.4 & 59.7 & 61.0 & 45.8 & 43.4 & 70.3 & 63.9 & 52.4 & 76.8 & 58.3 \\
%  & MDDA \cite{wang2020transfer} & 54.9 & 75.9 & 77.2 & 58.1 & 73.3 & 71.5 & 59.0 & 52.6 & 77.8 & 67.9 & 57.6 & 81.8 & 67.3 \\
 \cmidrule(){1-15}
 \multirow{3}*{Adv Based} & DANN \cite{ganin2015unsupervised} & 45.6 & 59.3 & 70.1 & 47.0 & 58.5 & 60.9 & 46.1 & 43.7 & 68.5 & 63.2 & 51.8 & 76.8 & 57.6 \\
%  & ADDA \cite{tzeng2017adversarial} &  \\
%  & CAN \cite{zhang2018collaborative} & \\
 & CDAN \cite{long2017conditional} & 46.6 & 65.9 & 73.4 & 55.7 & 62.7 & 64.2 & 51.8 & \textbf{49.1} & 74.5 & 68.2 & \textbf{56.9} & \textbf{80.7} & 62.8  \\
 & DDAN \cite{wang2020transfer} & \textbf{51.0} & 66.0 & 73.9 & 57.0 & 63.1 & 65.1 & 52.0 & 48.4 & 72.7 & 65.1 & 56.6 & 78.9 & 62.5 \\
 \midrule
% & DAMPC (Ours) & 46.53 & \textbf{68.42} & \textbf{75.24} & \textbf{58.30} & \textbf{66.30} & \textbf{67.48} & \textbf{56.74} & 44.77 & \textbf{75.33} & \textbf{69.26} & 51.94 & 80.33 & \textbf{63.39} \\
% & PC-NAS (Ours) & 46.19 & 66.03 & 73.70 & 57.89 & 63.48 & 65.80 & 56.94 & 44.19 & 73.58 & 69.02 & 51.11 & 78.89 & xx \\
& DAMPC-NAS (Ours) & 46.53 & \textbf{68.42} & \textbf{75.24} & \textbf{58.3} & \textbf{66.3} & \textbf{67.48} & \textbf{56.94} & 44.77 & \textbf{75.33} &	\textbf{69.26} & 51.94 & 80.33 & \textbf{63.4} \\

 \bottomrule
 \end{tabular}
 }
%  \vskip 0.1in
 \label{tab:officehome}
\end{table*}

\subsection{Setup}

We conduct experiments on three benchmark datasets, including Office-31 \cite{saenko2010adapting},  Office-Home \cite{venkateswara2017deep}, and VisDA-2017 \cite{peng2017visda}.
The Office-31 dataset has 4,652 images in 31 categories collected from three distinct domains: \textit{Amazon} (\textbf{A}), \textit{Webcam} (\textbf{W}) and \textit{DSLR} (\textbf{D}). We can construct six transfer tasks: \textbf{A} $\rightarrow$ \textbf{W}, \textbf{D} $\rightarrow$ \textbf{W}, \textbf{W} $\rightarrow$ \textbf{D}, \textbf{A} $\rightarrow$ \textbf{D}, \textbf{D} $\rightarrow$ \textbf{A}, and \textbf{W} $\rightarrow$ \textbf{A}.
The Office-Home dataset consists of 15,500 images in 65 object classes under the office and home settings, forming four extremely dissimilar domains: \textit{Artistic} (\textbf{Ar}), \textit{Clip Art} (\textbf{Cl}), \textit{Product} (\textbf{Pr}), and \textit{Real-World} (\textbf{Rw}) and 12 transfer tasks.
The VisDA-2017 dataset has over 280K images across 12 classes. It contains two very distinct domains: \textbf{Synthetic}, which contains renderings of 3D models from different angles and with different lightning conditions, and \textbf{Real} that are natural images. On this dataset, we study a transfer task: Synthetic $\rightarrow$ Real.

We compare the proposed \textbf{DAMPC-NAS} method with state-of-the-art DA methods, including Joint Distribution Adaptation (\textbf{JDA}) \cite{long2013transfer}, Deep Domain Confusion (\textbf{DDC}) \cite{tzeng2014deep}, Deep Adaptation Network (\textbf{DAN}) \cite{long2015learning}, Domain Adversarial Neural Network (\textbf{DANN}) \cite{ganin2015unsupervised}, Correlation Alignment for Deep Domain Adaptation (\textbf{D-CORAL}) \cite{sun2016deep}, Residual Transfer Networks (\textbf{RTN}) \cite{long2016unsupervised}, Joint Adaptation Networks (\textbf{JAN}) \cite{long2017deep},  Adversarial Discriminative Domain Adaptation (\textbf{ADDA}) \cite{tzeng2017adversarial}, Conditional Domain Adversarial Networks (\textbf{CDAN}) \cite{long2017conditional}, Collaborative and Adversarial Network (\textbf{CAN}) \cite{zhang2018collaborative}, Manifold Dynamic Distribution Adaptation (\textbf{MDDA}) \cite{wang2020transfer}, and Dynamic Distribution Adaptation Network (\textbf{DDAN}) \cite{wang2020transfer}. The results of baseline methods are directly reported from DDAN \cite{wang2020transfer} and CDAN \cite{long2017conditional}.

We use the PyTorch package \cite{paszke2017automatic} to implement all the models and leverage the ResNet-50 network \cite{he2016deep} pretrained on the ImageNet dataset \cite{russakovsky2015imagenet} as the backbone for the feature extraction. For optimization, we use the mini-batch SGD with the Nesterov momentum 0.9. The learning rate is adjusted by $\eta_p = \eta_0(1 + \alpha p)^{-\beta}$, where $p$ is the index of training steps, $\eta_0$ = 0.1, $\alpha$ = 0.001, and $\beta$ = 0.75. The batch size is set to 128 for all the datasets.

\subsection{Results}

\begin{figure*}[ht]
\centering
\includegraphics[width=0.9\textwidth]{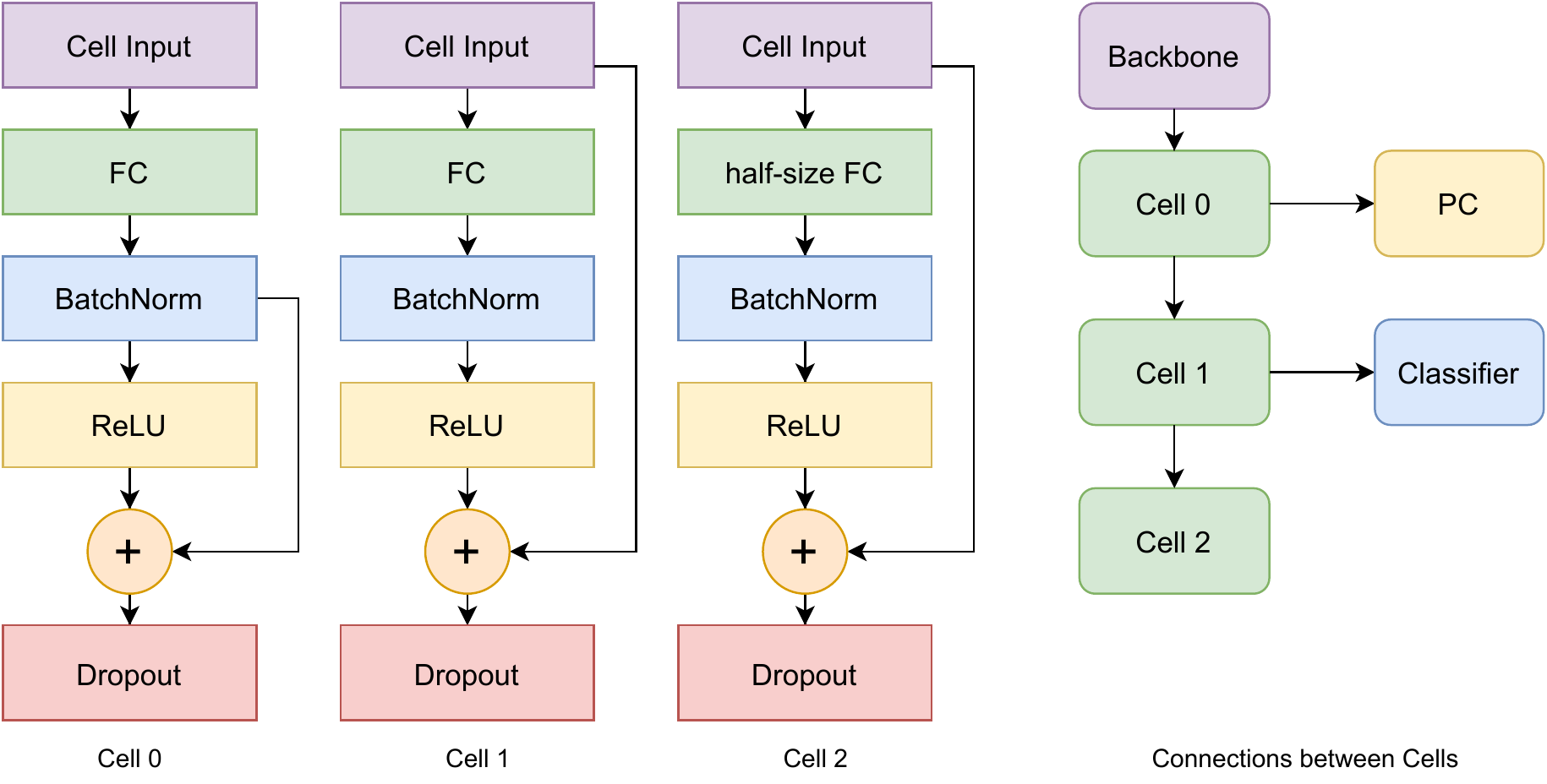}
\caption{Searched architecture for transfer task D$\rightarrow$W of the Office-31 dataset. Left: architectures within the three cells. Right: connections between the three cells, PC and classifier.}
\label{fig:arch_dw}
\end{figure*}

The classification results on the Office-31 dataset are shown in Table \ref{tab:office31}. As illustrated in Table \ref{tab:office31}, the proposed DAMPC-NAS method achieves the best average accuracy.
%On hard transfer tasks, e.g., D$\rightarrow$A and W$\rightarrow$A, where the source and target domains are substantially different and the source domain is smaller than the target domain, DAMPC-NAS is inferior to previous state-of-the-art methods.
In four out of six transfer tasks, DAMPC-NAS performs the best, especially on transfer tasks A$\rightarrow$D and A$\rightarrow$W, which is transferring from a large source domain to a small target domain and in the other two tasks, the DAMPC-NAS method performs slightly worse than the best baseline method, which implies that the proposed DAMPC-NAS model works well when the source data is sufficient and it is able to learn transferable feature representations for effective domain adaptation.

% These encouraging results highlight the key importance of maximizing population correlation in deep neural networks, and suggest that

Figure \ref{fig:arch_dw} shows the architecture found by DAMPC-NAS for the transfer task D$\rightarrow$W constructed on the Office-31 dataset. The left part of Figure \ref{fig:arch_dw} shows the search choice within the three cells found by the DAMPC-NAS method and the right part of Figure \ref{fig:arch_dw} shows the connections among the three cells, PC and classifier. In Cell 0, the DAMPC-NAS method chooses the FC layer with the same size as the input and the skip connection is connected to the batch-norm layer. In Cell 1, the choice of FC is the same as Cell 0 but the skip connection is starting from the cell input. In Cell 2, the skip connection is the same as Cell 2 but the FC layer is of half size of the input. For connections between cells, the DAMPC-NAS method chooses to use the output of Cell 0 to calculate the PC and the output of Cell 1 to calculate the classification loss. For a simple transfer task D$\rightarrow$W, the searched architecture only has two cells, which indicates that the DAMPC-NAS method can adaptively learn an architecture depending on the the complexity of the DA task. Moreover, the location of the skip connection moves forward in Cell 1 and Cell 2 when compared with Cell 0, which is to help reduce the network depth and alleviate the vanishing gradient problem.

Table \ref{tab:officehome} shows the classification results on the Office-Home dataset. According to the results, we can see that DAMPC-NAS achieves the best average accuracy and performs the best in eight out of twelve transfer tasks. while transferring from a large source domain to a small target domain (i.e., Cl$\rightarrow$Ar, Pr$\rightarrow$Ar, and Rw$\rightarrow$Ar), DAMPC-NAS achieves the best performance and this phenomenon is similar to the Office-31 dataset, which again demonstrate that the proposed DAMPC-NAS model works well when the source data is sufficient.

According to experimental results on the most challenging VisDA-2017 dataset as shown in Table \ref{tab:visda}, the proposed DAMPC-NAS method  outperforms all the baseline methods by improving by $1.9\%$ over state-of-the-art baseline methods (i.e., CDAN) on this dataset, which again demonstrates the effectiveness of the proposed method.

\begin{table}[!htbp]
\centering
\caption{Accuracy (\%) on the VisDA-2017 dataset with ResNet-50 as the backbone.}
\resizebox{\linewidth}{!}{
\begin{tabular}{llc}
\toprule
 Type & Method & Synthetic$\rightarrow$Real\\
 \midrule
 & ResNet-50 \cite{he2016deep} & 45.6 \\
 \cmidrule(){1-3}
%  \multirow{3}*{Dist Based} & JDA \cite{long2013transfer} &  \\
%  & DDC \cite{tzeng2014deep} &  \\
  \multirow{3}*{Dist Based} & DAN \cite{long2015learning} & 53.0 \\
 & RTN \cite{long2016unsupervised} & 53.6 \\
%  & D-CORAL \cite{sun2016deep} & \\
 & JAN \cite{long2017deep} & 61.6 \\
%  & MDDA \cite{wang2020transfer} &  \\
 \cmidrule(){1-3}
 \multirow{2}*{Adv Based} & DANN \cite{ganin2015unsupervised} & 55.0 \\
%  & ADDA \cite{tzeng2017adversarial} &  \\
 & CDAN \cite{long2017conditional} & 66.8 \\
%  & CAN \cite{zhang2018collaborative} & \\
%  & DDAN \cite{wang2020transfer} & \\
 \midrule
 & DAMPC-NAS (Ours) & \textbf{68.75} \\
%  & PC-NAS (Ours) & 68.32 \\
 \bottomrule
 \end{tabular}
 }
 \label{tab:visda}
\end{table}

\subsection{Ablation Study}

Firstly, we conduct an ablation study on the Office-31, Office-Home, and VisDA-2017 datasets to demonstrate the effectiveness of the proposed PC. We compare PC with widely used distance functions, including Proxy $\mathcal{A}$-distance, Kullback-Leibler divergence (KL-divergence), Maximum Mean Discrepancies (MMD),  CORrelation ALignmen (CORAL), and Central Moment Discrepancy (CMD). For fair comparison, we only replace the minus of the PC with these distance functions in Eq. (\ref{eq:obj}). Specifically, we adopt the ResNet-50 as the backbone, following with the bottleneck layer (consisting of a fully connected layer, a batch normalization layer, a ReLU activation function, and a dropout function) used for generating hidden features and a fully connected layer used for prediction. According to experimental results shown in Tables \ref{tab:ablation_study_office31}, \ref{tab:ablation_study_visda} and \ref{tab:ablation_study_officehome}, we can see that none of the distance functions can obtain performance improvement compared with no distance function used (i.e., ResNet-50). One possible reason is that the normalization layer used in the bottleneck layer has improved the performance of the ResNet-50 and adapting these distance functions can not improve the performance further. However, the proposed PC can still obtain performance improvement over ResNet-50, which indicates the effectiveness of the proposed PC.

\begin{table}[!htbp]
\centering
\caption{Ablation Study on the Office-31 dataset with ResNet-50 as the backbone.}
\resizebox{\linewidth}{!}{
\begin{tabular}{lccccccc}
\toprule
 Measurement & A$\rightarrow$D & A$\rightarrow$W & D$\rightarrow$A & D$\rightarrow$W & W$\rightarrow$A & W$\rightarrow$D & Avg \\
 \midrule
 None &  83.53 & 80.50 & 64.61 & 98.49 & 62.69 & 100.0 & 81.64 \\
 Proxy $\mathcal{A}$-distance &  82.73 & 81.01 & 64.04 & 98.11 & 61.77 & 100.0 & 81.28 \\
 KL-divergence &  83.94 & 79.75 & 63.90 & 97.86 & 63.51 & 99.80 & 81.46 \\
 MMD &  83.13 & 79.25 & 64.11 & 98.74 & 63.12 & 100.0 & 81.39 \\
 CORAL & 84.34 & 80.25  & 64.61 & 98.24 & 62.80 & 99.80 & 81.67 \\
 CMD & 82.93 & 79.50 & 64.29 & 98.62 & 63.10 & 100.0 & 81.41  \\
 \midrule
 PC (Ours) &  88.35 & 91.32 & 70.36 & 98.49 & 69.05 & 100.0 & 86.26  \\
%  PC-NAS (Ours) &  88.55 & 93.08 & 66.31 & 97.61 & 66.63 & 100.0 & 85.36 \\
 \bottomrule
 \end{tabular}
 }
 \label{tab:ablation_study_office31}
\end{table}

\begin{table}[!htbp]
\centering
\caption{Ablation Study on the VisDA-2017 dataset with ResNet-50 as the backbone.}
% \resizebox{\linewidth}{!}{
\begin{tabular}{lc}
\toprule
 Measurement & Synthetic$\rightarrow$Real\\
 \midrule
 None & 57.68 \\
 Proxy $\mathcal{A}$-distance & 56.36 \\
 KL-divergence & 56.27\\
 MMD & 58.76  \\
 CORAL  & 56.66  \\
 CMD & 56.65  \\
 \midrule
 PC (Ours) & 65.25 \\
 \bottomrule
 \end{tabular}
%  }
 \label{tab:ablation_study_visda}
\end{table}

\begin{table*}[ht]
%\vskip -0.1in
\centering
%\vspace{0.2cm}
\caption{Ablation Study on the Office-Home dataset with ResNet-50 as the backbone.}
\resizebox{\linewidth}{!}{
\begin{tabular}{lccccccccccccc}
\toprule
 Measurement & Ar$\rightarrow$Cl & Ar$\rightarrow$Pr & Ar$\rightarrow$Rw & Cl$\rightarrow$Ar & Cl$\rightarrow$Pr & Cl$\rightarrow$Rw & Pr$\rightarrow$Ar & Pr$\rightarrow$Cl & Pr$\rightarrow$Rw & Rw$\rightarrow$Ar & Rw$\rightarrow$Cl & Rw$\rightarrow$Pr & Avg  \\
 \midrule
 None & 43.41 & 66.55 & 74.64  & 56.61 & 63.98 & 65.32 & 53.36 & 39.36 & 72.64 & 64.73 & 46.30 & 76.55 & 60.29 \\
 Proxy $\mathcal{A}$-distance & 43.21 & 65.44 & 74.85 & 55.09 & 62.51 & 65.37 & 52.33 & 38.63 & 72.83 & 64.57 & 46.23 & 76.66 & 59.81 \\
 KL-divergence & 44.01 &  66.75 & 74.50 & 55.75 & 63.42 & 66.51 & 52.74 & 38.14 & 73.43 & 65.84 & 44.79 & 77.13 & 60.25 \\
 MMD & 43.78 &  66.28 & 74.48 & 55.62 & 64.07 & 66.19 & 53.40 & 38.30 & 73.15 & 64.89 & 45.52 & 77.43 & 60.26 \\
 CORAL & 44.15 & 65.85 & 74.16 & 55.42 & 63.01 & 66.83 & 52.95 & 39.38  & 72.53 & 65.14 & 45.96 & 77.07 & 60.20 \\
 CMD & 44.40 & 65.92 & 74.50  & 54.68 & 63.37 & 67.07 & 52.78 & 38.88  & 72.94 & 65.64 & 45.29 & 77.36 & 60.24 \\
 \midrule
 PC (Ours) & 46.19 & 66.03 & 73.7 & 57.89 & 63.48 & 65.80 & 56.94 & 44.19  & 75.58 & 69.02 & 51.11 & 78.89 & 62.24 \\
 \bottomrule
 \end{tabular}
 }
%  \vskip 0.1in
 \label{tab:ablation_study_officehome}
\end{table*}

\begin{figure*} [ht]
%\vskip -0.1in
  \centering
    \subfigure[ResNet-50]{\includegraphics[width=0.33\textwidth]{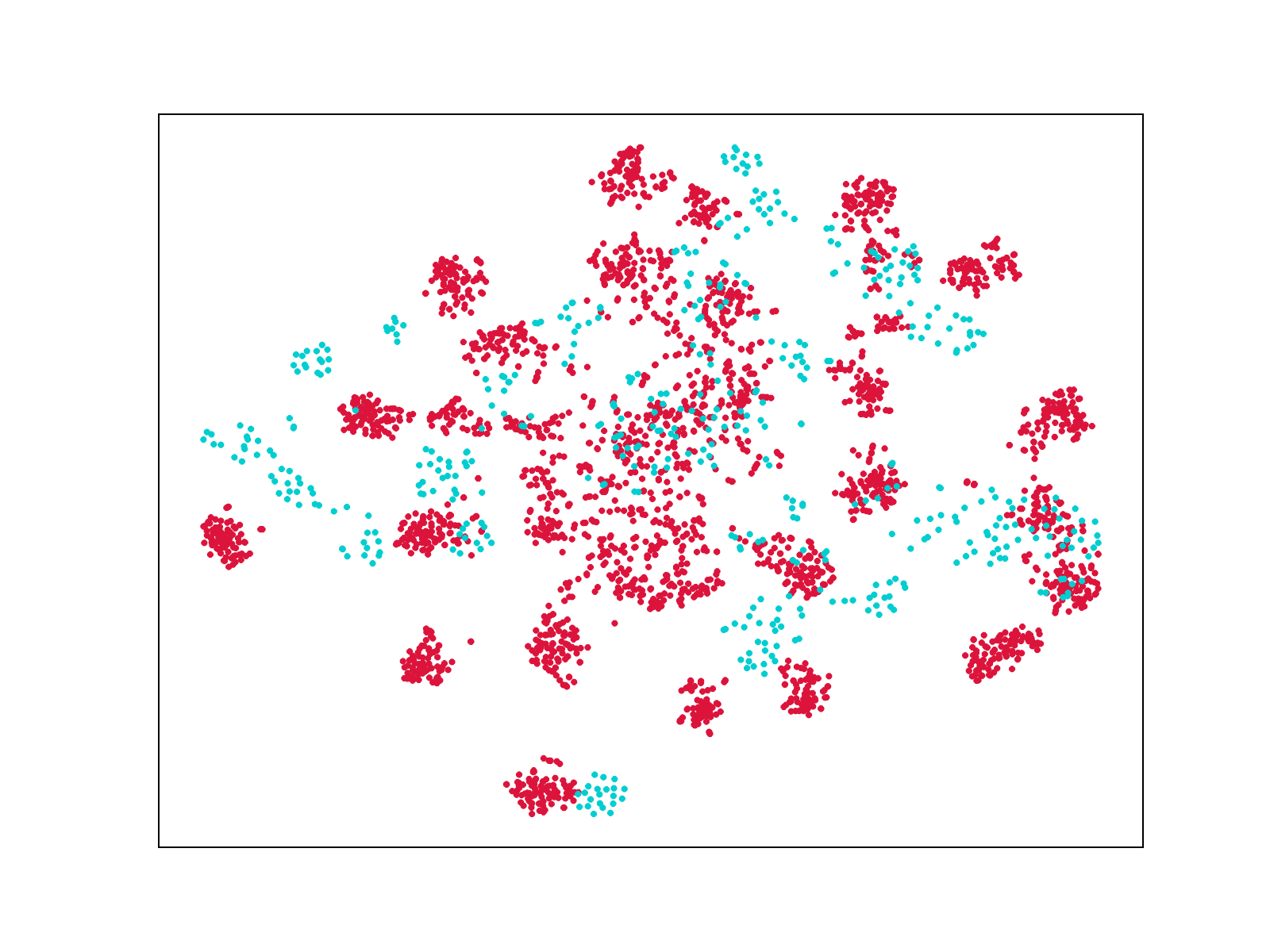}}
    \subfigure[DAN]{\includegraphics[width=0.33\textwidth]{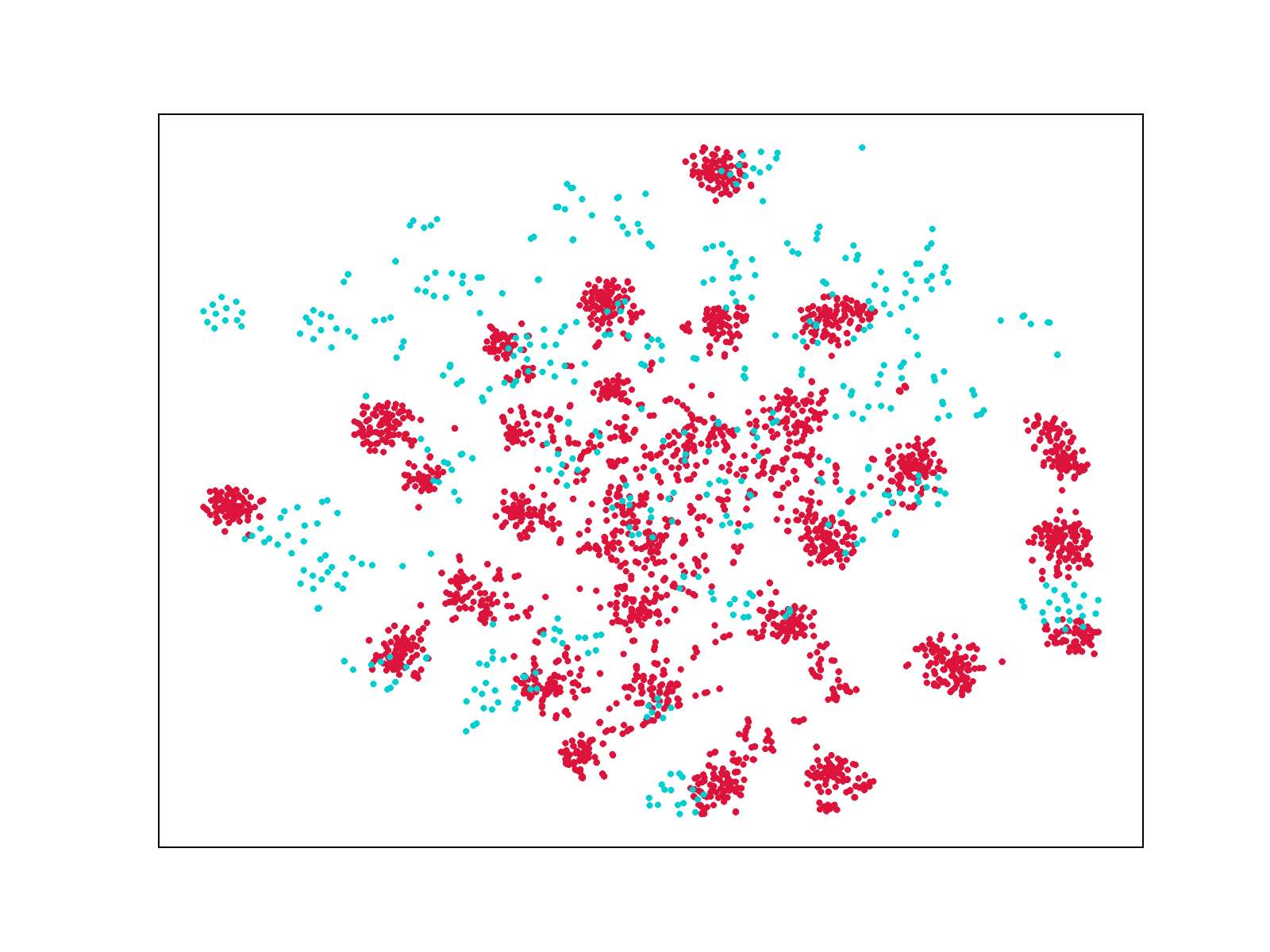}}
    \subfigure[DAMPC-NAS (Ours)] {\includegraphics[width=0.33\textwidth]{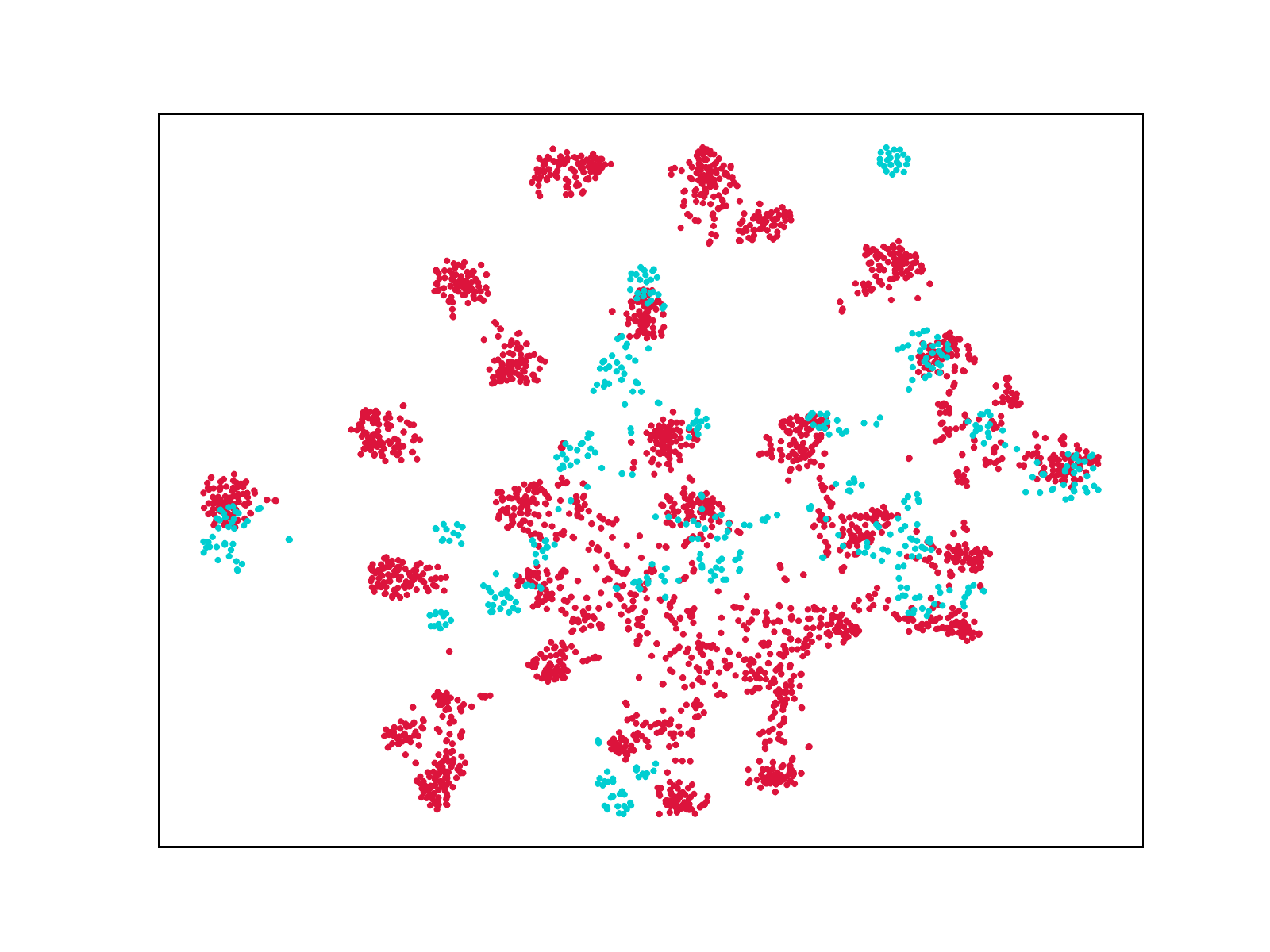}}
  \caption{t-SNE visualization of different methods for the transfer task A$\rightarrow$D in the Office-31 dataset.}
\label{fig:visualization}
\end{figure*}

\begin{figure}[!htbp]
\centering
\includegraphics[width=\linewidth]{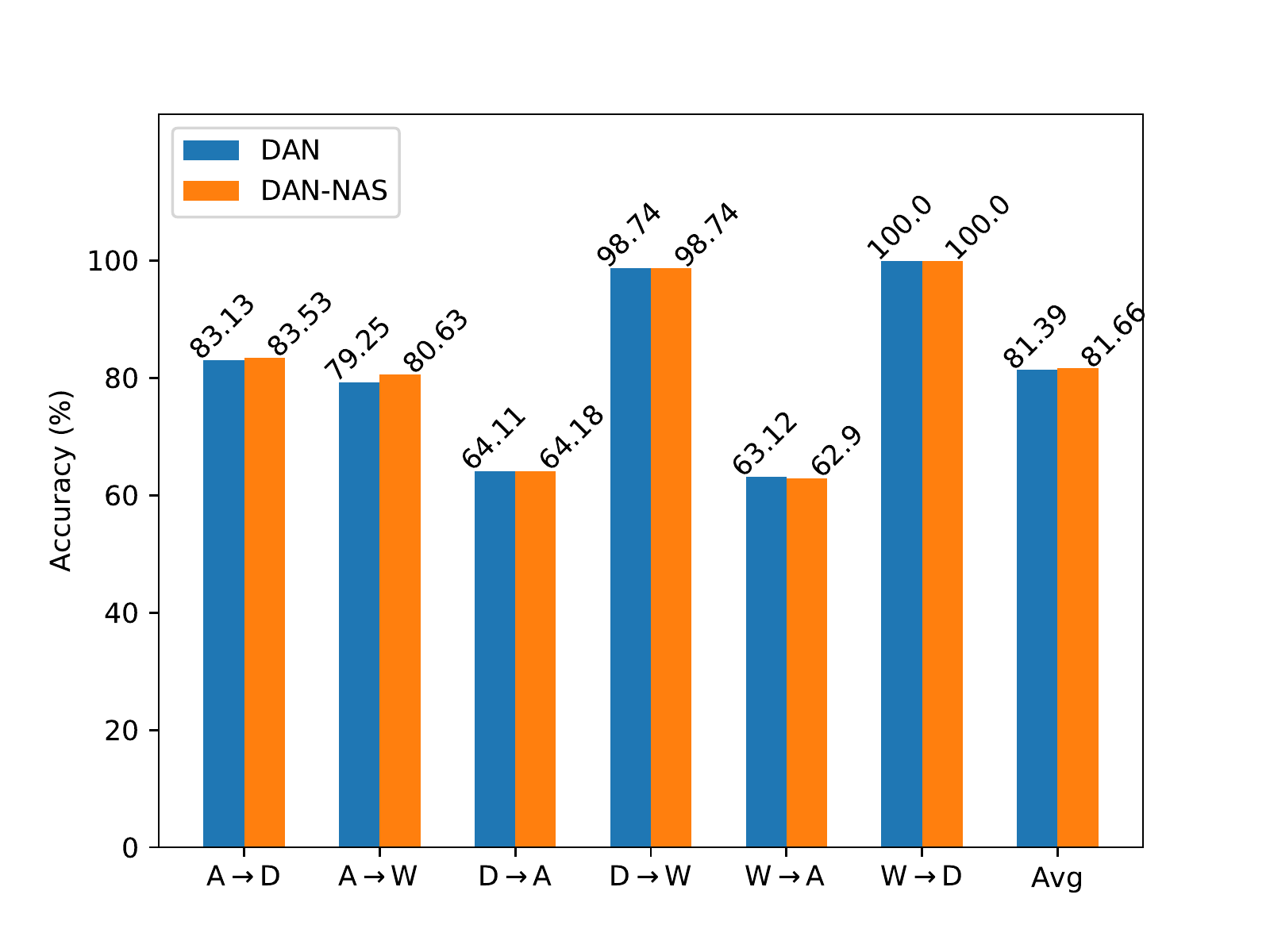}
\caption{DAMPC-NAS with DAN on the Office-31 dataset.}
\label{fig:mmd_nas}
\end{figure}

Then we conduct another ablation study on the Office-31 dataset to demonstrate the effectiveness of the architecture searching process in the DAMPC-NAS method. Specifically, we modify Algorithm \ref{alg:nas} to search an optimal architecture for the DAN by replacing the minus of the PC with MMD in $\mathcal{L}_{\text{DAMPC}}$. According to experimental results shown in Figure \ref{fig:mmd_nas}, DAN-NAS performs comparable to and even slightly better than DAN on the six transfer tasks in the Office-31 dataset, which demonstrates the usefulness of the search process in the DAMPC method.

\subsection{Visualization}

We visualize in Figure \ref{fig:visualization} the hidden feature representations of the transfer task A$\rightarrow$D constructed on the Office-31 dataset learned by ResNet-50 which is trained on source samples only, DAN, and DAMPC-NAS, respectively. According to Figure \ref{fig:visualization}, we can see that samples with the representations learned by ResNet-50 and DAN are not distinguishable, but those by DAMPC-NAS are more separable, which implies that the proposed DAMPC-NAS method can learn discriminative and transferable feature representations for DA.

\section{Conclusion}

In this paper, we propose a new DA method called DAMPC based on the proposed PC function that can measure the domain similarity. We further design the DAMPC-NAS framework that searches optimal network architectures for DA tasks. Experiments results on the Office-31, Office-Home, and VisDA-2017 datasets demonstrate the effectiveness of the proposed method. Moreover, the proposed DAMPC-NAS framework has shown its potential to search optimal architectures for other DA methods. In our future studies, we will apply the proposed  the DAMPC-NAS framework to search architectures for other DA methods and other DA settings.

\bibliography{NASUDA}
% \bibliographystyle{plain}

%\appendix

%\section{Appendix}

\end{document}